\documentclass[10pt,twocolumn,letterpaper]{article}

\usepackage{cvpr}
\usepackage{times}
\usepackage{epsfig}
\usepackage{graphicx}
\usepackage{amsmath}
\usepackage{amssymb}

\usepackage{color}

\usepackage[breaklinks=true,bookmarks=false]{hyperref}

\cvprfinalcopy 

\def\cvprPaperID{10039} 

\begin{document}

\title{Revisiting Pose-Normalization for Fine-Grained Few-Shot Recognition}

\author{Luming Tang\qquad Davis Wertheimer\qquad Bharath Hariharan\\
Cornell University\\
{\tt\small \{lt453,dww78,bh497\}@cornell.edu}
}

\maketitle

\begin{abstract}
Few-shot, fine-grained classification requires a model to learn subtle, fine-grained distinctions between different classes (e.g., birds) based on a few images alone. This requires a remarkable degree of invariance to pose, articulation and background. A solution is to use pose-normalized representations: first localize semantic parts in each image, and then describe images by characterizing the appearance of each part. While such representations are out of favor for fully supervised classification, we show that they are extremely effective for few-shot fine-grained classification. With a minimal increase in model capacity, pose normalization improves accuracy between 10 and 20 percentage points for shallow and deep architectures, generalizes better to new domains, and is effective for multiple few-shot algorithms and network backbones. Code is available at \href{https://github.com/Tsingularity/PoseNorm_Fewshot}{\url{https://github.com/Tsingularity/PoseNorm_Fewshot}}.
\end{abstract}

\section{Introduction}
\begin{figure*}
\centering
\includegraphics[width=0.9\textwidth]{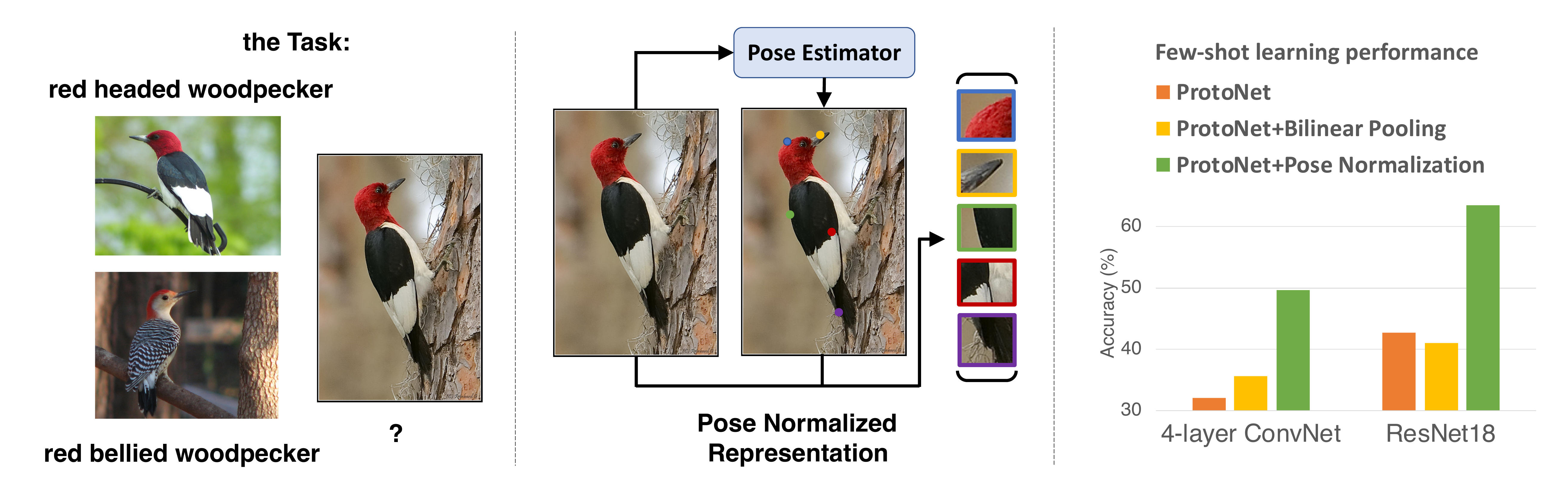}
\caption{\textbf{Left:} The fine-grained few-shot recognition task. Objects share the same part structure and differences between categories are subtle. \textbf{Middle:} Based on a simple pose estimator, a pose-normalized representation can capture semantic part information. \textbf{Right:} On both shallow and deep backbones, pose normalization increases few-shot learning performance significantly. A shallow architecture with our representation (4-layer ConvNet+Pose Normalization) even outperforms a much deeper blackbox network without it (ResNet18).}
\label{fig:intro}
\end{figure*}
The ability to generalize with minimal fine-tuning is a crucial property for learned neural models, not just to unseen data but also to unseen \textit{types} of data. 
Consider the task shown in Figure~\ref{fig:intro}. 
We are given just a single image (or a very small number) from a few bird species, and from this information alone we must learn to recognize them.
Humans are known to be very good at this \emph{few-shot learning} task~\cite{Schmidt2009}, but machines struggle: in spite of dramatic progress in visual recognition and two years of focused research,
performance on several few-shot benchmarks remains far below that of fully supervised approaches.

This is a problem in practice, especially for fine-grained classification problems (such as that in Figure~\ref{fig:intro}). In this setting, distinct classes can number in the hundreds, while the expertise and effort required to correctly label these classes can make annotation expensive. Together, this makes the collection of large labeled training sets for fine-grained classification difficult, sometimes prohibitively so. The ability of neural networks to handle fine-grained, few-shot learning can thus be crucial for real-world applications. 

What is the reason behind the large gap between machine and human performance on this task?
An intuitive hypothesis is that humans use a much more stable feature representation, which is invariant to large spatial deformations.
For example, in the bird classification task, we might characterize a bird image using the attributes of its various parts: the shape of the beak, the color of the wing, the presence or absence of a crown. 
Such a characterization is invariant not just to changes in the image background, but also to variation in camera pose and articulation, allowing us to effectively perceive similarities and differences across a wide range of bird species, and individual images of them.

Such a featurization is \emph{``pose normalized''}, and was explored as a promising direction for fine-grained classification before the re-discovery of convolutional networks~\cite{zhang2012pose}.
Researchers found, however, that end-to-end training with black-box architectures, and without pose normalization, led to great improvement in the standard benchmarks (albeit with consistent modifications, such as bilinear pooling~\cite{lin2015bilinear}).
Indeed, in recent years, winners on the annual fine-grained classification challenges~\cite{fgvc} have mostly focused on these black-box architectures.
The intuitive idea of pose normalization has fallen by the wayside.

In contrast, we argue that the dominance of black-box architectures over pose-normalized representations is an artifact of the fully-supervised classification problem.
In these settings, all classes to be distinguished are known \emph{a priori}, and we have significant amounts of training data for each class.
This reduces the need for pose and background invariance, since the training data will likely include a broad range of variation within each class. At the same time, leveraging category-specific biases in pose and background will likely be beneficial, since the representation need not generalize to new classes.
These factors act in favor of black-box architectures with no built-in inductive biases.
However, if we want the learnt model to \emph{adapt to new classes from limited data}, as in few-shot learning, the intuitive invariance of pose normalization becomes more useful.

In this paper, we revisit pose normalization for the task of few-shot, fine-grained classification, and demonstrate its usefulness in this setting.
Pose normalization is implemented through an extremely simple modification to convolutional architectures, adding very few new parameters (in contrast to prior methods that increase network size by a factor of two or higher~\cite{zhang2015fine, guo2019aligned}).
Our method is orthogonal to the choice of few-shot learning technique and backbone neural architecture.
We evaluate our approach on three different few-shot learning techniques, two differently-sized backbone architectures, and three fine-grained classification datasets of bird species  and aircraft. 
We find that:

\begin{enumerate}
\item Pose normalization provides significant gains across the board, in some cases providing a more than 20 point improvement in accuracy, while requiring \emph{no} part annotations for novel classes.
\item In all settings, pose normalization outperforms black-box modifications to the neural architecture, such as bilinear pooling. 
\item The advantages of pose normalization are apparent even when as little as only 5\% of the base class training data is annotated with pose.
\item Pose normalization is effective for both shallow and deep network architectures. Shallow networks with pose normalization \emph{outperform} deeper blackbox ones.
\end{enumerate}

The large performance gains we observe, along with the simplicity of the architecture itself, points to the power of pose normalization in fine-grained, few-shot classification.

\section{Related Work}
\textbf{Fine-grained recognition} is a classic problem in computer vision, and a recurring challenge~\cite{fgvc}.
While we focus on bird species classification~\cite{WahCUB_200_2011}, the presented ideas apply to other fine-grained tasks, such as identifying models of aircraft~\cite{maji13fine-grained}, cars~\cite{cars}, or any other problem where objects have a consistent set of parts.
In the context of fine-grained recognition, Farrell et al.~\cite{farrell2011birdlets} proposed the idea of pose normalization: predicting the parts of the object and recording the appearance of each part as a descriptor. 
Many versions of the idea have since been explored, including varying the kind of parts~\cite{zhang2012pose, guo2019aligned, zhang2014panda}, the part detector~\cite{zhang2014part}, and the combination of these ideas with neural networks~\cite{zhang2015fine}.
The last of these is the most similar to our work.
However, all of these approaches are concerned with fully supervised recognition, whereas here we look at few-shot recognition.

Pose normalization has also served as inspiration for black-box models where the parts are unsupervised.
Lin et al.~\cite{lin2015bilinear} introduce bilinear pooling as a generalization of such normalization, and we compare to this idea in our work.
Spatial Transformer Networks~\cite{jaderberg2015spatial} instantiate unsupervised pose normalization explicitly and train it end-to-end. Other instantiations of this intuition have also been proposed \cite{dai2017deformable,han2019p,sun2019learning}. However, these unsupervised approaches add significant complexity and computation, making it difficult to discern the benefits of pose-normalization alone. In contrast, we focus on a lightweight, straightforward, semantic  approach to show that pose normalization, not added network power, is responsible for improved performance.

\textbf{Few-shot learning} methods can be loosely organized into the following three groups: 1) Transfer learning baselines train standard classification networks on base classes, and then learn a new linear classifier for the novel classes on the frozen representation.
Recent work has shown this to be competitive~\cite{chen2019closerfewshot,wang2019simpleshot,nakamura2019revisiting}. 2) Meta-learning techniques train a ``learner'': a function that maps small labeled training sets and test images to test predictions. Examples include ProtoNet~\cite{snell2017prototypical}, MatchingNet~\cite{vinyals2016matching}, RelationNet~\cite{sung2018learning} and MAML~\cite{finn2017model}. 
These learners might sometimes include learnt data augmentation~\cite{wang2018low}, which some methods train using pose annotations~\cite{DixitCVPR2017}.
3) Weight generation techniques generate classification weights for new categories~\cite{gidaris2018dynamic, gidaris2019generating}.

Most few-shot learning methods use blackbox network architectures, which function well given enough labeled data, but may suffer in the highly constrained few-shot learning scenario.
Wertheimer and Hariharan~\cite{WertheimerCVPR2019} revisit the bilinear pooling of Lin et al.~\cite{lin2015bilinear} and find it to work well. They also introduce a simple, effective localization-normalized representation, but which is limited to coarse object bounding boxes instead of fine-grained parts. Zhu et al.~\cite{zhu2019learning} introduce a semantic-guided multi-attention module to help zero-shot learning, but is fully unsupervised. We compare to an unsupervised baseline in our experiments.

Pose normalization increases invariance to common modes of variation.
An alternative to increasing invariance is to use learnt data augmentation~\cite{hariharan2017low, wang2018low, DixitCVPR2017}. However, this typically requires large additional networks and significant computation. Instead, we focus on a lightweight approach.
Note also that one of our baselines~\cite{gidaris2018dynamic} already outperforms a recent augmentation method~\cite{wang2018low}. 

In the following sections, we first overview few-shot recognition. We then show that pose-normalization of features can act as a plug-and-play network layer in a range of few-shot learning algorithms.

\section{Few-Shot Recognition}
The goal of few-shot learning is to build a \emph{learner} that can produce an effective classifier given only a small labeled set of examples. 
In the classic few-shot setting, the learner is first provided a large labeled set (the \emph{representation set}, $D_{repre}$) consisting of many labeled images from base classes $Y_{base}$.
The learner must set its parameters, and any hyper-parameters, using this data.
It then encounters a disjoint set of novel classes $Y_{novel}$ from which it gets a small set of reference images $D_{refer}$.
The learner must then learn a classifier for the novel classes from this set.

In most techniques, we can divide the learner into three modules: a feature-map extractor $f_\theta$, a feature aggregator $g_\phi$, and a learning algorithm $h_w$.

\textbf{The feature map extractor $f_\theta$} is usually implemented as a deep convolutional neural network, with learnable parameters $\theta$. For each input image $x$, the network yields the corresponding feature map tensor $\boldsymbol{F}{=}f_\theta (x){\in} \mathbb{R}^{C{\times}H{\times}W}$, where $C,H,W$ denote respectively the channel, height, and width dimensions of the feature map.

\textbf{The feature aggregator $g_\phi$} is a transformation parameterized by $\phi$, converting feature maps into global feature vectors: $\boldsymbol{v}{=}g_\phi(\boldsymbol{F}){\in} \mathbb{R}^d$, where $d$ is the latent dimensionality.
Typically $g_\phi$ is a global average pooling module.

\textbf{The learning algorithm $h_w$} takes a dataset $S$ of training feature vectors and corresponding labels, and a test feature vector $\boldsymbol{v}$, and outputs a probability distribution over labels $\hat{p}$ for the latter:
 $\hat{p}(x){=}h_w(\boldsymbol{v}, S)$. For our purposes we consider three representative methods:

\textit{Transfer learning} follows the standard network pretraining and fine-tuning procedure. $h_w$ is implemented by a simple linear classifier with a learned weight matrix and softmax activation. Functions $f_\theta,g_\phi$ are trained concurrently with $h_w$, minimizing the standard cross-entropy loss over data in $D_{repre}$. To adapt the model to novel classes, feature extractor parameters $\theta,\phi$ are frozen, and $h_w$ trains a new linear classifier on the novel classes in $D_{refer}$.

\textit{Prototypical network}~\cite{snell2017prototypical} is a representative meta-learning method that produces a prototype representation for each class by averaging the feature vectors within that class. $h_w$ is then a non-parametric classifier assigning class probabilities based on the distance between a datapoint's feature vector and each class prototype. Every training episode samples N classes from the base categories $Y_{base}$, and a small support set and query set of images from within each one. Support images form class prototypes, while N-way classification on the query set produces the loss, and corresponding update gradients to parameters $\theta,\phi$. 

In \textit{Dynamic few-shot learning}~\cite{gidaris2018dynamic}, $h_w$ is once again a linear (or cosine) classifier, but instead of being directly fine-tuned on $D_{refer}$, the classifier is generated by a learnt \emph{weight generator} $G$. The training process consists of two stages. The first is standard classification training on $D_{repre}$. During the second stage, the feature extractor parameters $\theta,\phi$ are frozen. To train the generator $G$, the algorithm randomly picks several ``fake'' novel classes from $Y_{base}$, and treats them as if they were truly novel, performing classification with the classifier weights generated by $G$ and minimizing the classification loss on simulated ``test'' examples from these classes. 

\section{Pose-Normalized Feature Vectors}
\label{sec:model}
\begin{figure*}
\centering
\includegraphics[width=0.9\textwidth]{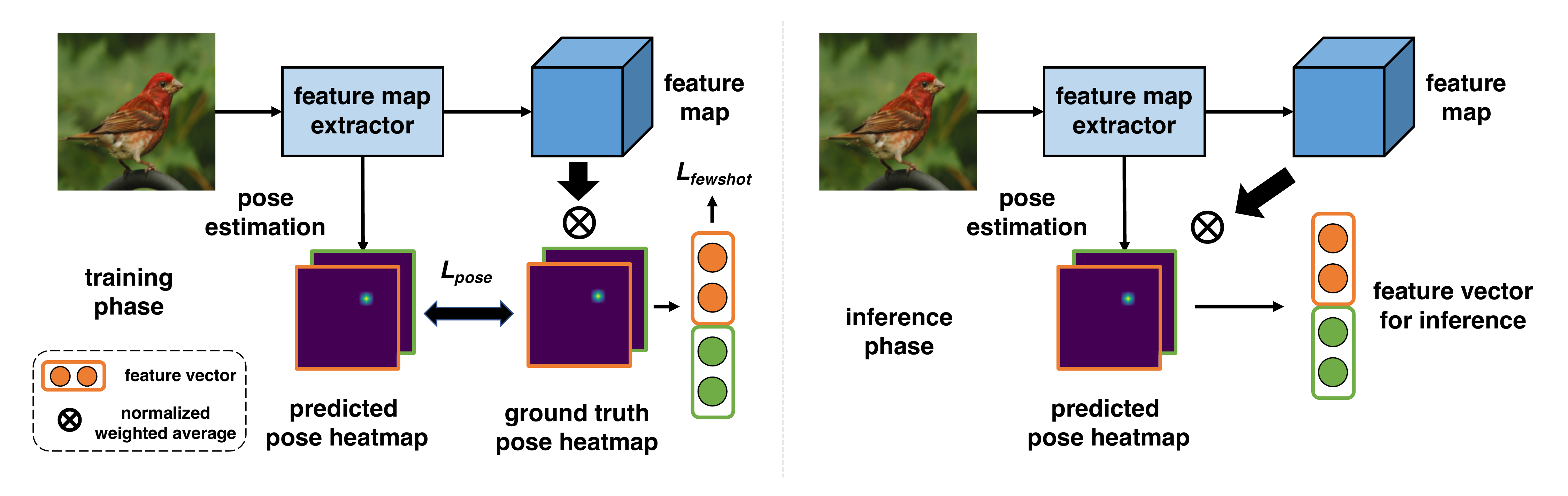}
\caption{The pose normalization framework for training and inference. The pose estimator takes an intermediate output of the network backbone as input and generates pose heatmap predictions. The feature vector is calculated by applying each heatmap as an attention over the feature map. The final representation is the concatenation of these vectors. In this example, the number of parts $M{=}2$.}
\label{fig:model}
\end{figure*}
Two intuitions motivate our proposed method. First, for fine-grained recognition, the difference in appearances between two classes tends to be extremely small. In the few-shot setting, it is even harder for an algorithm to capture these subtle differences, as only a few examples are available for reference. Using pose normalization to focus the feature representation on the most informative parts of each image should then benefit the learning process. Second, because fine-grained recognition involves similar kinds of objects, they are likely to share the same semantic structures. 

Thus it is highly probable that a pose estimator trained on base classes will generalize, even to unseen novel classes.

We assume $M$ distinct parts. 
Part annotations are available for (some) base-class training samples in $D_{repre}$, but \emph{not for novel classes}. We format part annotations for each image $x$ as an $M{\times}H{\times}W$ location tensor $\boldsymbol{m^*}$, where $H{\times}W$ is the spatial resolution of the feature map.

We now present our method for extracting pose-normalized feature vectors.
For this, the network must first estimate pose.
We use an extremely small, two-layer convolutional network $q_\phi$.
This operates on a feature map tensor $\boldsymbol{F}'{\in}\mathbb{R}^{C'{\times}H'{\times}W'}$ extracted from an intermediate layer of the feature map extractor $f_\theta$.
$q_\phi$ uses sigmoid activation in the final layer and produces a heatmap location prediction for all annotated parts $\boldsymbol{m} {=}q_\phi(\boldsymbol{F}'){\in}\mathbb{R}^{M{\times}H{\times}W}$. 
We deliberately use a small $q_\phi$ and \emph{reuse} computation in $f_\theta$ to minimize the effect the additional parameters might have on the final performance of the classifier. Improved performance should indicate that \emph{pose information} is useful for fine-grained few-shot learning, not a larger network. 

Given the heatmap $\boldsymbol{m}$ and feature map $\boldsymbol{F}$, we must construct a feature vector $\boldsymbol{v}$. Each channel in $\boldsymbol{m}$ is applied as a spatial attention mask to the feature map, producing an attention-normalized feature vector. Concatenating these $M$ part description vectors produces the final representation vector for the image. Formally, denoting $\boldsymbol{F}(h,w){\in}\mathbb{R}^C$ as the feature vector for location $(h,w)$ in feature map $\boldsymbol{F}$, and $m_i(h,w){\in}\mathbb{R}$ as the heatmap pixel value at position $(h,w)$ for the $i$-th part category, $\boldsymbol{v}{\in}\mathbb{R}^{CM}$ is calculated as:
\begin{align}
    \boldsymbol{v}_{i}&=\frac{\sum_{h,w}^{H,W}\boldsymbol{F}(h,w)\cdot m_i(h,w)}{\epsilon+\sum_{h,w}^{H,W}m_i(h,w)}\\
    \boldsymbol{v} &= [\boldsymbol{v}_1,\dots,\boldsymbol{v}_i,\dots,\boldsymbol{v}_{M}]
\end{align}
where $\epsilon{=}10^{-5}$. The loss during training is the sum of the pixel-wise log loss between the ground truth part location heatmap $\boldsymbol{m^*}$ and the predicted heatmap $\boldsymbol{m}$, and the original few-shot classification loss:
\begin{align}
\label{eq:loss}
    L_{pose} &= -\frac{1}{MHW}\sum_{i,h,w}^{M,H,W} [m^*_i(h,w)\log m_i(h,w) \nonumber\\
    &+(1-m^*_i(h,w))\log(1-m_i(h,w))]\\
    L_{total} &= L_{fewshot}+\alpha\cdot L_{pose}
\end{align}

where $\alpha$ is a balancing hyper-parameter. To facilitate learning in the classification branch, feature vectors for few-shot classification are initially produced from the ground truth part annotation heatmap $\boldsymbol{m^*}$ instead of the predicted heatmap $\boldsymbol{m}$. Afterwards, the pose estimation network's parameters $\phi$ are frozen. In subsequent adaptation/fine-tuning and evaluation/inference stages on novel classes, feature vectors are calculated from the predicted heatmap $\boldsymbol{m}$. An overview of our approach is provided in figure~\ref{fig:model}.

Note that while we assume a fixed set of consistent part labels during training, we do not require parts to consistently appear across all objects, nor must any particular object contain all the specified parts. Thus, our pose estimator should generalize broadly: \emph{any} fine-grained classification of objects that depends on the appearance of various parts (e.g., cars, furniture, insects) is amenable to this approach.

\section{Experiments}
\subsection{Datasets and implementation details}
\label{bird}
We experiment with the \textbf{CUB dataset}~\cite{WahCUB_200_2011} which consists of 11,788 images from 200 classes. It also includes 15 part annotations for each image, thus $M{=}15$. Following the evaluation setup in~\cite{WertheimerCVPR2019,chen2019closerfewshot}, we randomly split the dataset into 100 base, 50 validation and 50 novel classes. Base category images form the representation set $D^{CUB}_{repre}$. For each validation and novel class, we randomly sample 20\% of its images to form the reference set $D^{CUB}_{refer}$. The remaining novel images form the query set $D^{CUB}_{query}$, which is used for evaluating algorithms. \emph{Note that our models have access to part annotations only in base classes}. No part annotation information is available for any image in the validation or novel classes, including both their reference and query sets.

\textbf{NABird evaluation}: There are only 50 novel classes in CUB's evaluation set, which can potentially make evaluation noisy. The accuracy differences between few-shot learning algorithms also decrease significantly in the presence of domain shift~\cite{chen2019closerfewshot}. Thus, in order to verify the robustness and generalization capacity of our proposed method, we also evaluate our CUB models on another, much larger bird dataset: NABird~\cite{van2015building} (NA), which, after removing overlap with CUB,  contains 418 classes and 35,733 images. As before, we randomly sample 20\% of images from each category to form the reference set $D^{NA}_{refer}$. The remaining images form the query set $D^{NA}_{query}$.

\textbf{Network backbone}: For the feature map extractor $f_\theta$, previous work~\cite{snell2017prototypical,WertheimerCVPR2019,gidaris2018dynamic} adopts a standard architecture: a 4-layer, 64-channel convolution network with batch normalization and ReLU. In this setting, the input image size is $84{\times}84$ and the output feature map is $64{\times}10{\times}10$. Deeper backbones can significantly reduce the differences in performance between these methods~\cite{chen2019closerfewshot}, so in addition to the 4-layer network, we also train and evaluate a ResNet18~\cite{he2016deep} backbone, with a few technical modifications that increase performance across all models. We change the stride of the last block's first convolution and downsampling layers from 2 to 1. The output size of the last block thus remains at $14{\times}14$ instead of $7{\times}7$. We also add a $1{\times}1$ convolution with batch normalization to the last layer of the original ResNet18, which reduces the number of channels from 512 to 32. The input size of our modified ResNet18 is still $224{\times}224$, but the output size becomes $32{\times}14{\times}14$.

\textbf{Pose estimation module}: The layers of the pose estimation network $q_\phi$ are composed as Conv-BN-ReLU-Conv, where Conv denotes $3{\times}3$ convolution. In the 4-layer ConvNet, $q_\phi$ takes as input the feature map after the second convolution. The number of input/output channels for the two convolution layers in $q_\phi$ are $64/30$ and $30/M$ where $M$ is the number of part categories. In the ResNet18, $q_\phi$ takes the third block's feature map as input, and the corresponding convolution channel sizes are $256/64$ and $64/M$. It can be seen that the number of learnable parameters introduced by $q_\phi$ is small compared to the original backbone network.

\subsection{Baseline methods}

For the few-shot learning algorithm, we denote transfer learning, prototypical networks, and dynamic few-shot learning as transfer, proto, and dynamic, respectively. We compare our proposed pose normalization approach (PN) with the following feature aggregation methods, across all learning algorithms and network backbones:

\textbf{Average pooling} is the most straightforward method, commonly adopted in previous work. All subsequent models use average pooling when a feature aggregator is not otherwise specified.

We also present a baseline that trains this average-pooled feature extractor and classifier jointly with a localizer, with the latter discarded at test time.
This \textbf{Multi-Task} model, denoted MT, examines whether pose estimation functions purely as a regularizer in few-shot training.

\textbf{Bilinear pooling} (BP)~\cite{lin2015bilinear} is an effective module for expanding the latent feature space and increasing expressive power in fine-grained visual classifiers. Recent work~\cite{WertheimerCVPR2019} found that BP can be adapted to prototypical networks, improving performance without increasing parameter count.

\textbf{Few-shot localization} (FSL)~\cite{WertheimerCVPR2019} uses bounding box annotations in the representation and reference sets. The model learns to localize an object before classifying it, thus improving few-shot classification accuracy. Since this model's localizer is learnt in a prototypical way, it doesn't introduce any additional convolutional layers. 

\textbf{Bounding box normalization} (bbN) is a more direct comparison to bounding box based methods that does not require box annotations for novel classes. We use the PN model but set $M{=}2$, and train the localizer to separate images into foreground/background regions based on the ground truth bounding boxes for base class training data. 

\textbf{Unsupervised pose normalization} (uPN) is based on unsupervised localization~\cite{WertheimerCVPR2019}, a competitive localization method where feature maps are partitioned into soft regions based on feature distance from a set of learned parameter vectors. 
Following the same core idea, we introduce $M{=}15$ learned, category-agnostic pose vectors, and spatially partition the feature map based on relative feature distance to each vector at each location. 
We mean-pool over the resulting 15 soft regions, as if they were 15 predicted part locations, to produce a feature vector for the classifier.
The pose vectors are learned parameters, trained end-to-end and jointly with the classifier architecture, requiring no part annotations or separate localization module.

In addition, we include an \textbf{oracle} version of our model:
\textbf{Pose normalization with ground truth pose} (PN\_gt). 

\subsection{Few-shot recognition results}
We first train all models on $D^{CUB}_{repre}$, using the validation set to select the best hyper-parameters and stopping point for each model. We then evaluate them on $D^{CUB}_{query}$ using the limited set of labeled novel class images in $D^{CUB}_{refer}$. For the evaluation metric, we use the all-way evaluation~\cite{WertheimerCVPR2019, hariharan2017low, wang2018low} rather than the commonly adopted 5-way task. The algorithm is required to distinguish all novel classes simultaneously, a more challenging setup. For the number of reference images, we consider both the standard 1-shot/5-shot and the all-shot setting proposed by~\cite{WertheimerCVPR2019}, i.e. utilizing all the labeled images for each novel category in $D^{CUB}_{refer}$.

\begin{figure}
\centering
\includegraphics[width=0.4\textwidth]{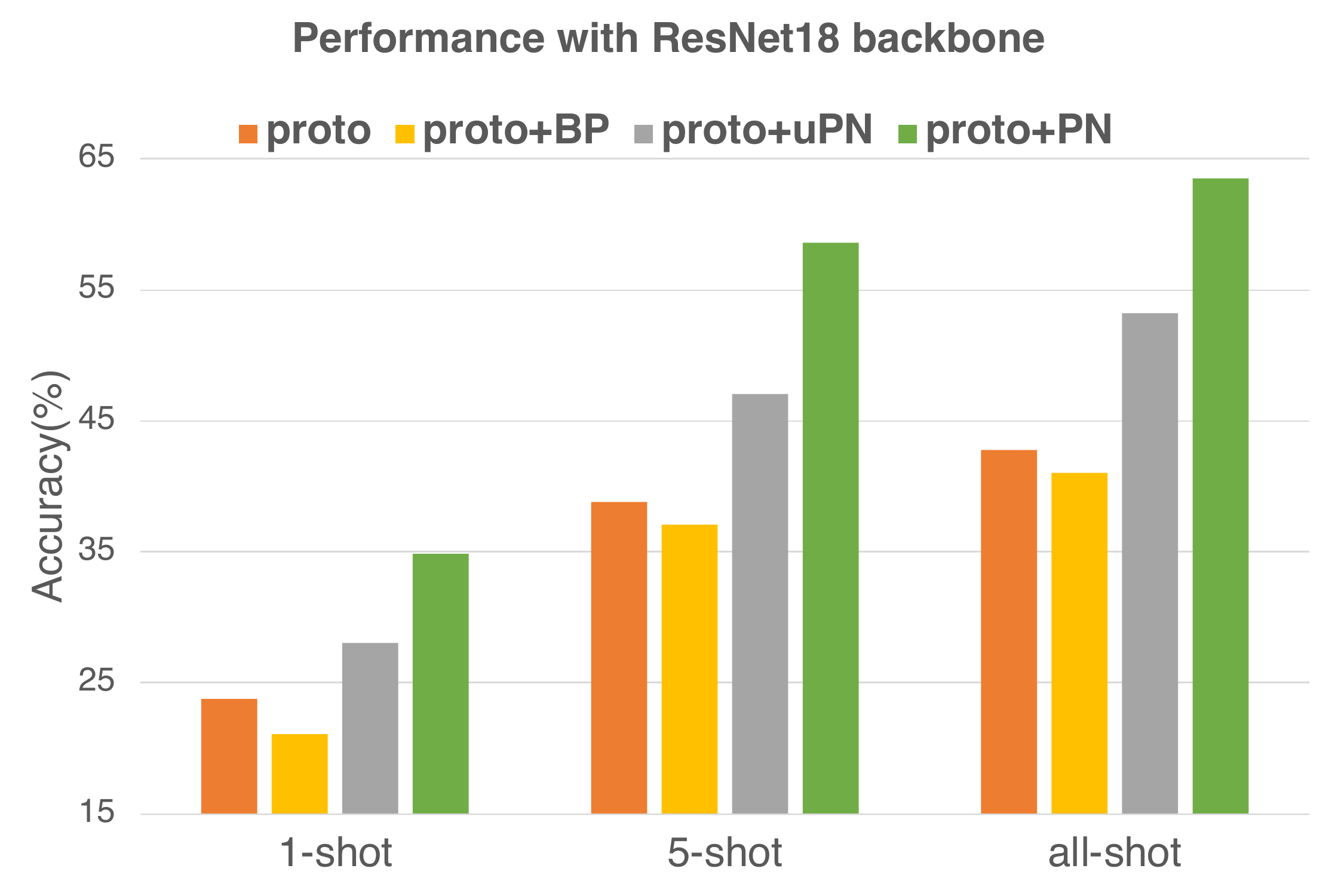}
\caption{Accuracy comparison on CUB. All models use a ResNet18 prototypical network. Pose normalization dominates other methods under all settings.}
\label{fig:shots_proto}
\end{figure}

\begin{figure}
\centering
\includegraphics[width=0.4\textwidth]{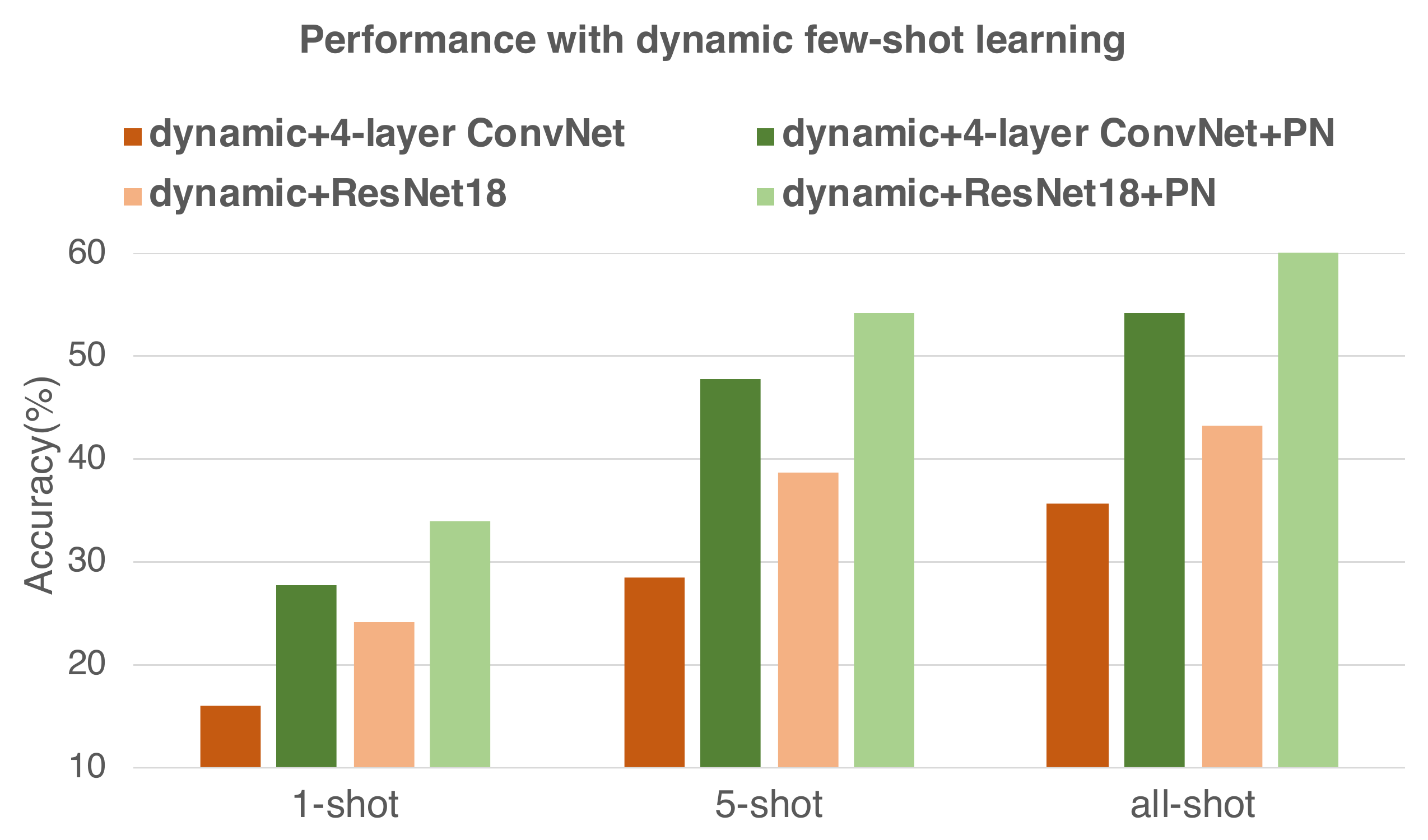}
\caption{Performance comparison for dynamic few-shot learning models on CUB. The accuracy boost from pose normalization is significant and consistent.}
\label{fig:shots_dynamic}
\end{figure}

For CUB, all-shot results are shown in table \ref{tab:cub}. For 1 and 5 shot settings, we plot the mean of 600 randomly generated test episodes in figures \ref{fig:shots_proto} and \ref{fig:shots_dynamic}. The 95\% confidence intervals are all less than 0.6 percentage points. 
Using the above models trained on CUB, we then do the same evaluation on NA, using $D^{NA}_{refer}$ and $D^{NA}_{query}$. The number of novel classes in NA is large (418), and the number of images per class is unbalanced. We therefore only report the all-way all-shot results in table \ref{tab:na}, with both mean accuracy over all test samples and mean accuracy per class. We average over 8 trials for the proto, proto+uPN, and proto+PN models in each of the above mentioned settings. 95\% confidence intervals are all within 0.9 percentage points.

From these experimental results, we conclude that:
\begin{enumerate}
\item \textbf{Pose normalization provides significant and consistent performance gains over the (average-pooled) baseline.} Accuracy improves for both shallow and deep network backbones, for all three few-shot learning approaches, and for both evaluation datasets. Under the all-way, all-shot setting on CUB, the accuracy gain is consistently greater than \textbf{15 points} for the 4-layer ConvNet, across all three learning algorithms, and reaches \textbf{20} points on ResNet18. Shallow networks with pose normalization can even outperform their deeper counterparts.
\item In all settings, \textbf{pose normalization outperforms other aggregation functions}, including black-box modifications (bilinear pooling), techniques based on bounding box localization (FSL and bbN) and unsupervised pose normalization.
It also outperforms multi-task training, indicating that normalization, rather than the additional auxiliary loss, is key. 

\item \textbf{Pose information is more effective than coarse object location.} In table \ref{tab:cub}, PN and bbN contribute similar quantities of new learnable parameters, but the fine-grained pose information in PN causes it to outperform bbN, which only focuses on a coarse bounding box. By comparing PN with PN\_gt, we see that a better pose estimator could potentially contribute an even larger boost to performance.
\end{enumerate}

\begin{table}
\centering
\resizebox{.38\textwidth}{!}{
\scriptsize
\setlength\tabcolsep{5pt}
\hskip-.02\textwidth
\begin{tabular} {  l  c  c}
\hline
\textbf{Model} & \textbf{4-layer ConvNet} & \textbf{ResNet18} \\
\hline
transfer & 33.42  & 46.47  \\
\textbf{transfer+PN} & \textbf{49.96}  & \textbf{57.53}  \\
\textit{transfer+PN\_gt} & 56.40 & 58.54  \\
\hline
proto & 32.09  & 42.73  \\
proto+MT & 35.56 & 50.93 \\
proto+BP & 35.56 & 41.04 \\
proto+FSL & 39.60  & 47.43 \\
proto+bbN & 37.75 & 44.02\\
proto+uPN & 46.24 & 53.18  \\
\textbf{proto+PN} & \textbf{49.56} & \textbf{63.44}  \\
\textit{proto+PN\_gt} & 59.55  & 62.63 \\
\hline
dynamic & 35.77 & 43.27  \\
\textbf{dynamic+PN} & \textbf{54.17}  & \textbf{60.19}  \\
\textit{dynamic+PN\_gt} & 62.67 & 60.09 \\
\hline
\end{tabular}
}
\vspace{.8mm}
\caption{Few-shot classification results for different models on the CUB dataset. Models are organized by few-shot learning algorithm, then by feature representation method. Pose normalization gives a significant performance boost for all three few-shot learning algorithms, with both shallow and deep network backbones.}
\label{tab:cub}
\end{table}

\begin{table}
\centering
\resizebox{.45\textwidth}{!}{
\scriptsize
\setlength\tabcolsep{5pt}
\hskip-.02\textwidth
\begin{tabular} {  l  c c c c}
\hline
\-\ & \multicolumn{2}{c}{\textbf{4-layer ConvNet}} & \multicolumn{2}{c}{\textbf{ResNet18}} \\
\textbf{Model} & \textbf{mean} & \textbf{per-class} & \textbf{mean} & \textbf{per-class}\\
\hline
transfer & 12.63  & 11.24  & 20.22  & 17.54\\
\textbf{transfer+PN} & \textbf{24.60}  & \textbf{21.76} & \textbf{28.36}  &\textbf{25.57} \\
\hline
proto & 8.73 & 8.37 & 13.33  & 12.55  \\
proto+MT & 10.59 & 10.10 & 16.41 &15.42 \\
proto+BP & 10.47  & 9.83 & 15.09  & 14.04  \\
proto+FSL & 12.34  & 11.61  & 15.62  & 14.81 \\
proto+bbN & 10.57  & 10.00 & 13.05  & 12.32  \\
proto+uPN & 18.91  & 17.51 & 22.12 & 20.77  \\
\textbf{proto+PN} & \textbf{21.02} & \textbf{19.47}  & \textbf{32.66}  & \textbf{30.59} \\
\hline
dynamic & 12.13  & 11.26  & 14.82  & 13.44  \\
\textbf{dynamic+PN} & \textbf{26.17} & \textbf{24.07} & \textbf{30.10}  & \textbf{27.86} \\
\hline
\end{tabular}
}
\vspace{.8mm}
\caption{Performance of CUB models on NA. The performance boost introduced by pose normalization is still significant in this new domain. Performance is consistent with CUB observations.}
\label{tab:na}
\end{table}

\subsection{Impact of the number of pose annotations}
\label{ablation}

While part locations are often cheaper to obtain than fine-grained expert class labels (see the careful labelling pipeline of~\cite{van2015building}), it could still be the case that high-quantity part annotations are difficult to collect.
We therefore consider an ablation of our model, where a limited number of training images have part annotations.
For the remaining images, $L_{pose}$ is not computed, and the predicted pose heatmap produces feature vectors for classifier training instead of the ground truth. 

\begin{figure}
\centering
\includegraphics[width=0.4\textwidth]{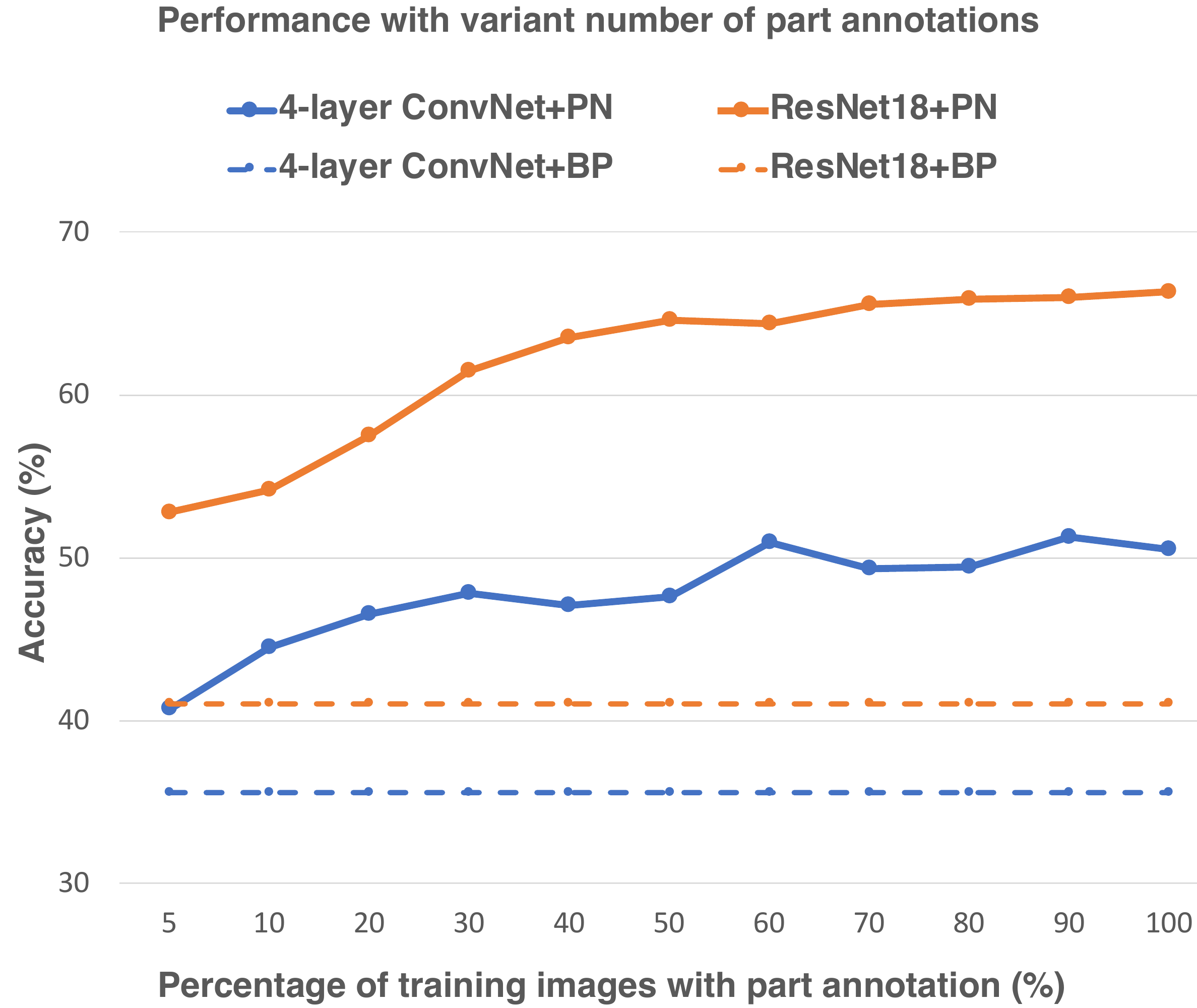}
\vspace{.8mm}
\caption{Few-shot test accuracy for pose normalization when part annotations are sparse. The performance drop is surprisingly small. Pose normalization outperforms bilinear pooling even when only 5\% of annotations are available during training.}
\label{fig:annotation}
\end{figure}

We evaluate prototypical networks on CUB with both shallow and deep backbones, and vary the percentage of images with part annotation. Results are given in figure \ref{fig:annotation}. Pose normalization is highly robust to annotation sparsity during training (less than 5 points fluctuation when above 30\% availability), and consistently outperforms BP even with as few as 5\% pose annotations available.

\subsection{Evaluation on FGVC-Aircraft}

We evaluate the generality of these conclusions on fine-grained aircraft classification~\cite{maji13fine-grained} (FGVC-Aircraft), which contains 10,000 images spanning 100 aircraft models. Following the same ratio as CUB, we split the classes into 50 base, 25 validation and 25 novel. The reference/query split is as described in Section \ref{bird}. Since this dataset doesn't contain any part annotation, we use an independent dataset OID-Aircraft~\cite{mahendran14understanding} (OID) to jointly train our pose normalization module. OID contains 6,357 images, ignoring those shared with FGVC, and 5 part annotations per image (thus $M{=}5$). OID contains no classification labels.

Each training iteration samples an image batch from OID and FGVC. OID images are used to calculate $L_{pose}$, while FGVC images use predicted pose heatmaps to get feature vectors. Results are shown in Table \ref{tab:fgvc}. Although the pose estimator is trained on disjoint images, it remains effective at boosting aircraft recognition performance. We conclude that pose normalization generalizes across fine-grained few-shot classification tasks. Extending this approach to non-fine-grained tasks or class-specific parts is not straightforward, but could be a valuable direction for future research.

\begin{table}
\centering
\resizebox{.5\textwidth}{!}{
\scriptsize
\setlength\tabcolsep{5pt}
\hskip-.02\textwidth
\begin{tabular} {  l  c c c c c c}
\hline
\-\ & \multicolumn{3}{c}{\textbf{4-layer ConvNet}} & \multicolumn{3}{c}{\textbf{ResNet18}} \\
\textbf{Model} & \textbf{1-shot} & \textbf{5-shot} & \textbf{all-shot} &  \textbf{1-shot} & \textbf{5-shot} & \textbf{all-shot}\\
\hline
proto & 24.40 & 43.24 & 52.06  & 46.27 & 63.15 & 67.76  \\
\textbf{proto+PN} & \textbf{26.04} & \textbf{50.35}  & \textbf{60.83}  & \textbf{58.72} & \textbf{77.75} &\textbf{81.96} \\
\hline
\end{tabular}
}
\vspace{.8mm}
\caption{Few-shot results under all three evaluation settings on the FGVC-Aircraft dataset. Results averaged over five trials.}
\label{tab:fgvc}
\end{table}

\section{Analysis}

\subsection{Model interpretation}
\label{interpretation}
\begin{figure*}
\centering
\includegraphics[width=0.85\textwidth]{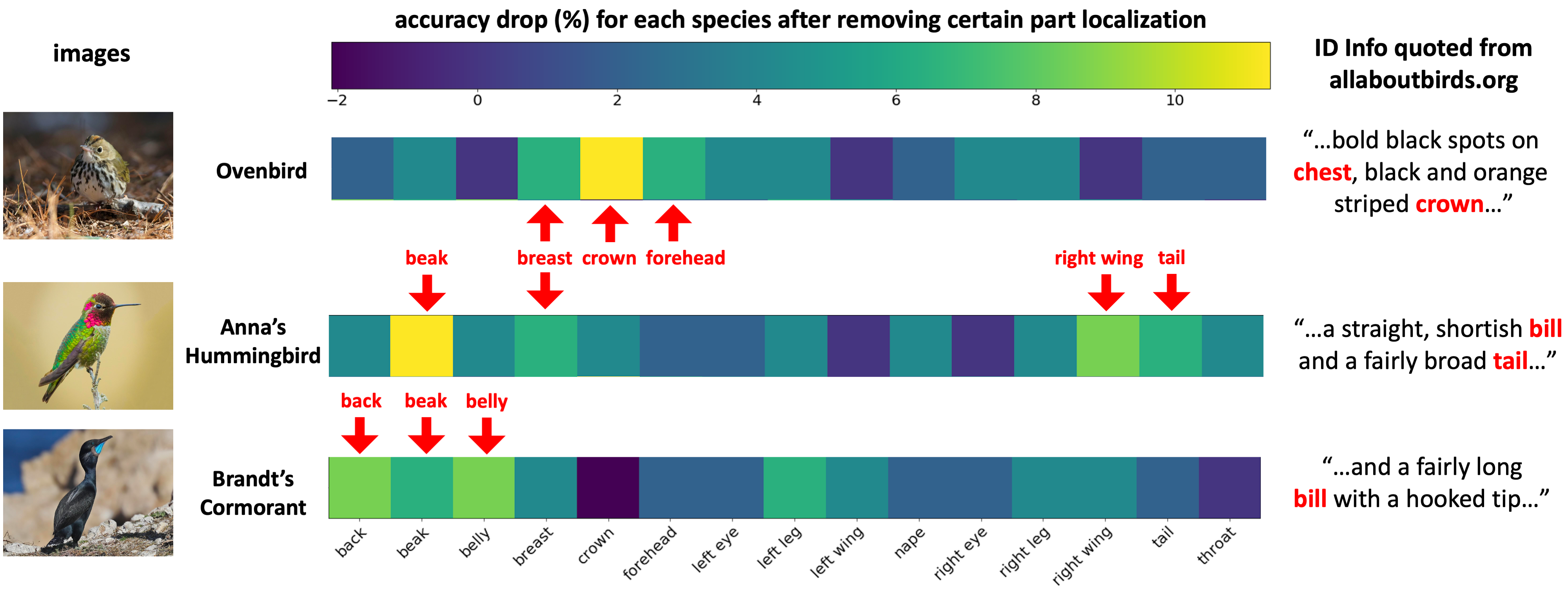}
\caption{ Visualizing accuracy drop for selected bird species when removing individual part vectors (part importance). 
On the right are quoted descriptions from bird experts on how to recognize those species. 
The estimated part importance matches well to expert judgments.}
\label{fig:matrix}
\end{figure*}

Accuracy notwithstanding, we would like for pose-normalized representations to be human-interpretable, unlike prior black-box representations. To investigate what the model actually learns, we conduct two experiments to analyze the learnt pose normalized representation. Both use the proto+PN model with a ResNet18 backbone.

\textbf{Part importance}: Every type of bird is likely to have a set of particularly distinguishable part attributes. To verify that our model learns this, we conduct the following test. For each class, we iterate over the parts and calculate the test accuracy when the corresponding part feature vector is removed from the representation.
The magnitude of the resulting drop in accuracy can be construed as the \emph{importance} of each part for this class as learned by the model.  
We visualize this learned importance for three species in figure \ref{fig:matrix} and compare it with species descriptions from a field guide.
Our network scores largely conform to expert judgments. 

\begin{figure}
\centering
\includegraphics[width=0.45\textwidth]{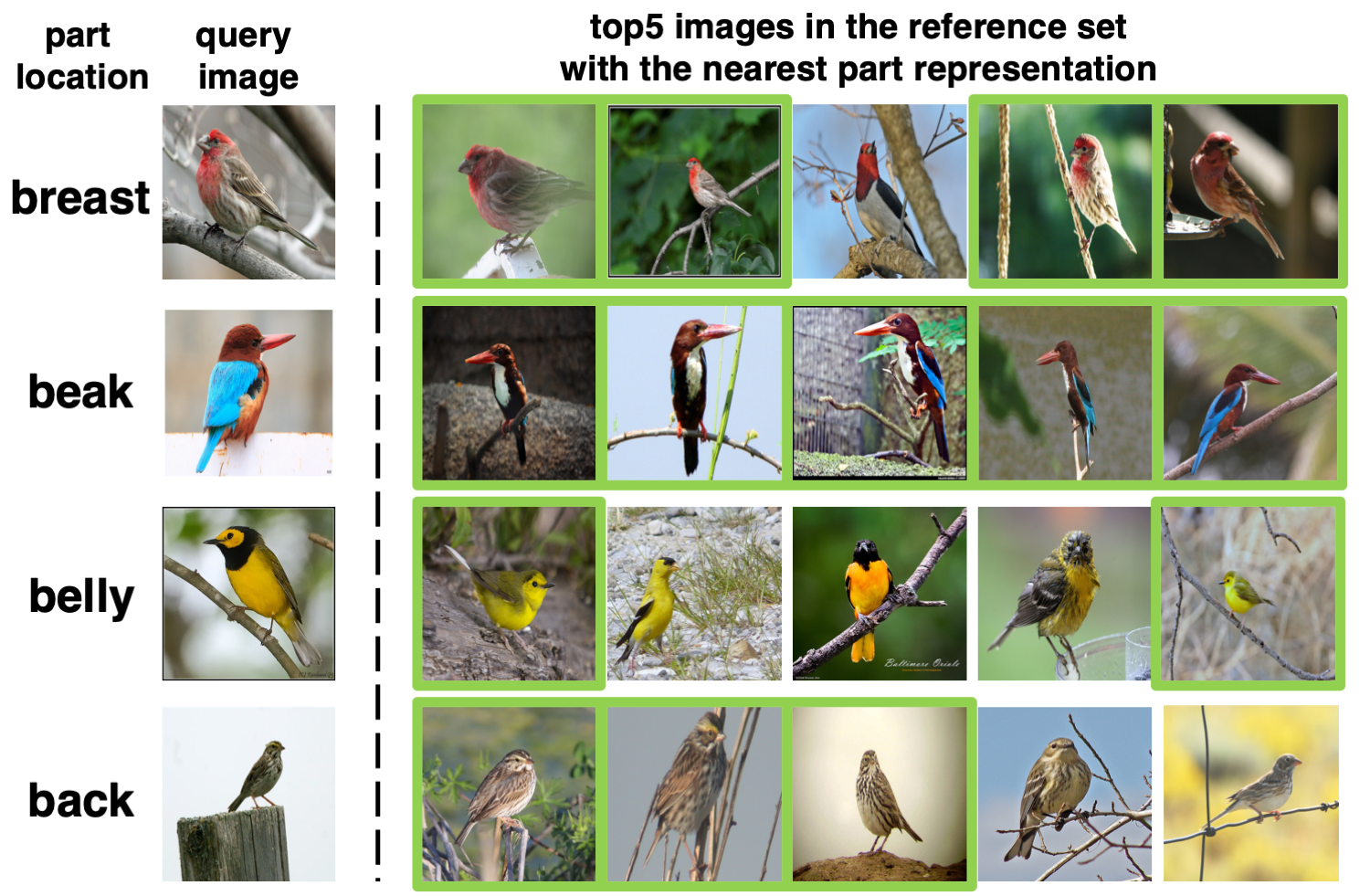}
\caption{Images with the closest part vector to the query image, for a given part location. Images are labeled with a green box if it belongs to the same category as the query image. We see that part representations capture semantically meaningful attributes of the part location across classes. }
\label{fig:neighbor}
\end{figure}

\textbf{Nearest neighbors}: Different birds might share the same part attribute; for example, the California Gull and the Ring-billed Gull have the same beak shape. Therefore in pose normalization, the beak vectors for these two birds should be close, as part vectors are designed to encode regional information in a class-agnostic way. To verify this, we find the top-5 images in the reference set with the closest part vectors to a given vector from a query image/part pair. Four random examples are given in figure \ref{fig:neighbor}. Generally, our assumption holds - the vector describing the given part in each query image does generalize to other species.

\subsection{Pose estimation}

\begin{figure}
\centering
\includegraphics[width=0.4\textwidth]{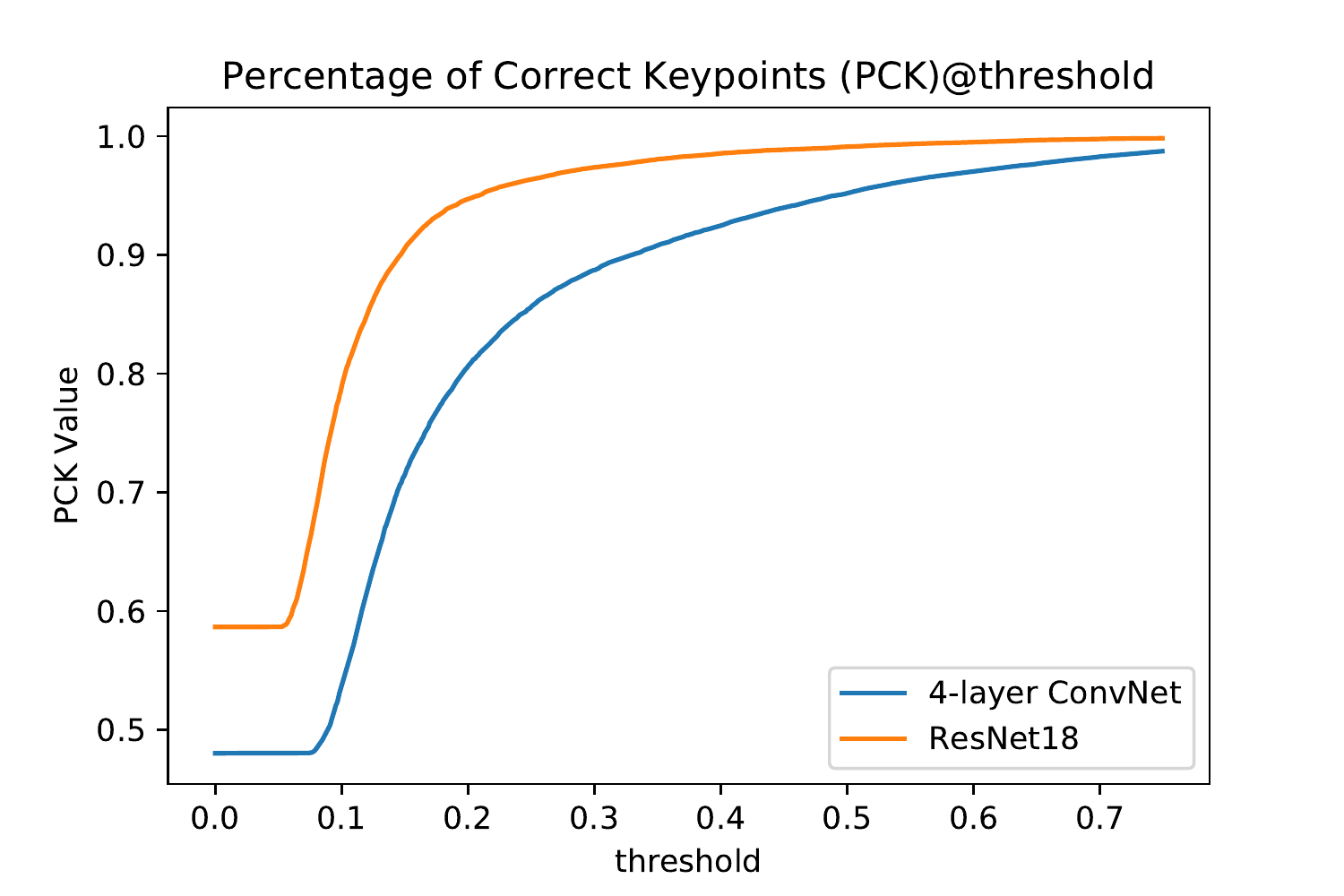}
\caption{Pose estimation results for different network backbones.}
\label{fig:pck}
\end{figure}

Following prior work on evaluating pose estimation~\cite{yang2012articulated,andriluka20142d} we calculate the normalized PCK  (normalized using the diagonal of the bounding box) at different thresholds for both shallow and deep network backbones. Results are given in figure \ref{fig:pck}. We see that both estimates can give accurate results. While the deeper network backbone does produce a better estimate, this boost is also quite limited. 
We believe that a more sophisticated pose estimator could lead to better results on few-shot recognition.  

\subsection{Unsupervised pose normalization}

\begin{figure}
\centering
\includegraphics[width=0.4\textwidth]{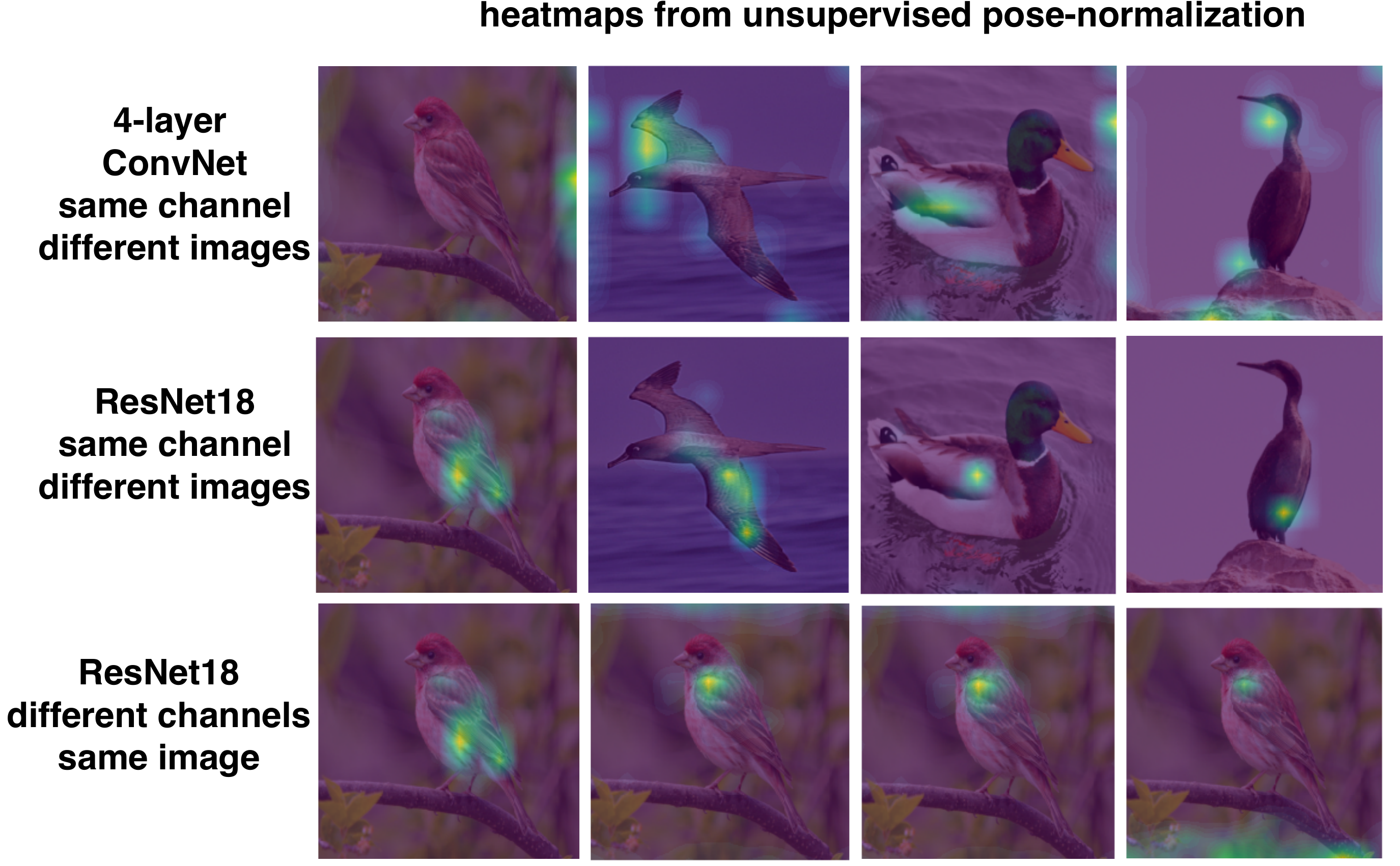}
\caption{Visualization of unsupervised heatmaps. Semantic content is highly inconsistent, and difficult to interpret meaningfully.}
\label{fig:unpart}
\end{figure}

We note that unsupervised pose normalization also performs well from a classification perspective (see table \ref{tab:cub}).
As shown in Figure \ref{fig:unpart}, the deeper backbone with unsupervised pose normalization does produce localized keypoints, which might help classification.
However, observe that the semantic meaning of these keypoints is not consistent across images (figure \ref{fig:unpart}, top two rows).
The prediction is also unstable, with different channels sometimes providing similar heatmaps (figure \ref{fig:unpart}, bottom row). 
This inconsistency could help to explain why machine-discovered parts underperform hand-designed ones in fine-grained, few-shot classification.

\section{Conclusion}
We show that a simple, lightweight pose normalization module can lead to consistently large performance gains on fine-grained few-shot recognition without any part annotations at test time. Our results hold for shallow and deep network backbones, multiple few-shot learning algorithms, and multiple domains. In addition to significant accuracy improvements, we also show that pose-normalized representations are highly human-interpretable. We therefore highly recommend pose normalization as a general area for the fine-grained few-shot learning community to revisit.

\section*{Acknowledgements}
\noindent This work was partly supported by a DARPA LwLL grant.
{\small
\bibliographystyle{ieee_fullname}
\bibliography{egbib}
}

\newpage
\onecolumn

\begin{center}
    \LARGE
    \textbf{Supplementary Materials}
\end{center}

\section{Training Details}
\label{details}

\textbf{Dataset splits}: The CUB and FGVC-Aircraft datasets contain class lists in a canonical order (found in classes.txt and variants.txt respectively). Using this ordering we assign each class an id starting from 1 for CUB and 0 for FGVC respectively. For each dataset, we split the categories by id into $50\%$ base, $25\%$ validation and $25\%$ novel classes following the rule: if id ${\bmod} 2{=}0$, id ${\in}$ base, else if id ${\bmod} 4{=}1$, id ${\in}$ validation, otherwise id ${\in}$ novel.

\textbf{Hyper-parameters for CUB}: For each model, we use the validation set to select the best hyper-parameters as well as the best stopping iteration during the training process. Here we give all the hyper-parameters finally chosen for training on CUB. All SGD optimizers use a default 0.9 momentum. PN models involve an additional hyper-parameter, the multiplier $\alpha$ for $L_{pose}$ (see formula \ref{eq:loss} in section \ref{sec:model}). This value is chosen as 100 for the 4-layer ConvNet and 200 for ResNet18.

For prototypical network based methods, we follow the same meta-training technique and batch sampling as \cite{WertheimerCVPR2019} and set each episode to 20-way 5-shot classification with 15 query images per class. The only exception is proto+FSL, for which we follow the same setting as \cite{WertheimerCVPR2019} and make each class in the episode consist of 5 refer shots for aggregating fore/background vectors, 5 shots for training for 10 shots for testing. Here we define one epoch as one pass over the whole representation set. The whole training process can then be divided into $s$ stages, each stage containing $e$ epochs, and the model is trained with optimizer $o$ and weight decay $\beta$, using an initial learning rate of $lr$ and cut by multiplying a factor $\gamma$ after finishing each stage. These hyper-parameters are shown in table \ref{tab:cub_hyper}. In addition, proto+bbN chooses $\alpha{=}10$ and proto+MT choose the same $\alpha$ value as proto+PN. During the whole training process, we evaluate the model on validation set every 20 epochs and select the best one to make the final evaluation on test set.

For transfer learning based methods, following the same notation as above, we list in table \ref{tab:cub_hyper} the hyper-parameters used for pre-training on base classes with batch size 64. For the finetuning on novel classes, we finetune the new classifier for 40 epochs with Adam, using learning rate of 0.001 and batch size of 16.

For dynamic few-shot learning based methods, we list in table \ref{tab:cub_hyper} the hyper-parameters used for the first training stage with batch size 64. For the second training stage, in each batch, we random sample 16 fake novel and 4 base classes, each class containing 20 images. We train the weight generator for 200 epochs with Adam, using learning rate 0.001. During the second training stage, we evaluate the model on validation set every 20 epochs and select the best one to make the final evaluation on test set.

\begin{table}[h]
\centering
\resizebox{.95\textwidth}{!}{
\scriptsize
\setlength\tabcolsep{1.5pt}
\hskip-.02\textwidth
\begin{tabular} {  l  c c c c c c | c c c c c c}
\hline
\-\ & \multicolumn{6}{c}{\textbf{4-layer ConvNet}} & \multicolumn{6}{c}{\textbf{ResNet18}} \\
\hline
\textbf{Model} & \textbf{optimizer} & \textbf{lr} & \textbf{$\gamma$} &  \textbf{epoch} & \textbf{stage} & \textbf{weight decay} & \textbf{optimizer} & \textbf{lr} & \textbf{$\gamma$} &  \textbf{epoch} & \textbf{stage} & \textbf{weight decay} \\
\hline
transfer & SGD & 0.1 & 0.1 & 200 & 2 & 5e-4 & SGD & 0.1 & 0.1 & 100 & 2 & 1e-3 \\
transfer+PN & SGD & 0.1 & 0.1 & 200 & 2 & 5e-4 & SGD & 0.1 & 0.1 & 100 & 2 & 1e-3 \\
transfer+PN\_gt & SGD & 0.1 & 0.1 & 200 & 2 & 5e-4 & SGD & 0.1 & 0.1 & 100 & 2 & 1e-3 \\
\hline
proto & SGD & 0.1 & 0.1 & 400 & 2 & 5e-4 & SGD & 0.1 & 0.1 & 300 & 2 & 1e-3 \\
proto+MT & SGD & 0.1 & 0.1 & 600 & 2 & 1e-3 & SGD & 0.1 & 0.1 & 300 & 2 & 5e-3 \\
proto+BP & Adam & 0.001 & NA  & 800 & 1 & 0 & Adam & 0.001 & NA & 600 & 1 & 1e-3  \\
proto+FSL & SGD & 0.01 & 0.1  & 400 & 2 & 5e-4 & SGD & 0.1 & 0.1 & 300 & 2 & 1e-3 \\
proto+bbN & SGD & 0.01 & 0.1 & 400 & 2 & 5e-4 & Adam & 0.1 & 0.5 & 160 & 5 & 0  \\
proto+uPN & SGD & 0.1 & 0.1 & 600 & 2 & 1e-3 & SGD & 0.1 & 0.1 & 200 & 2 & 5e-3  \\
proto+PN & SGD & 0.1 & 0.1 & 600 & 2 & 1e-3 & SGD & 0.1 & 0.1 & 300 & 2 & 5e-3 \\
proto+PN\_gt & SGD & 0.1 & 0.1 & 400 & 2 & 5e-4 & SGD & 0.1 & 0.1 & 300 & 2 & 5e-3 \\
\hline
dynamic & SGD & 0.1 & 0.1 & 200 & 2 & 5e-4 & SGD & 0.1 & 0.1 & 100 & 2 & 1e-3 \\
dynamic+PN & SGD & 0.1 & 0.1 & 100 & 2 & 5e-4 & SGD & 0.1 & 0.1 & 25 & 3 & 1e-3 \\
dynamic+PN\_gt & SGD & 0.1 & 0.1 & 50 & 2 & 5e-4 & SGD & 0.1 & 0.1 & 25 & 3 & 1e-3 \\
\hline
\end{tabular}
}
\vspace{.8mm}
\caption{The hyper-parameters selected for training on CUB.}
\label{tab:cub_hyper}
\end{table}

\textbf{Hyper-parameters for FGVC-Aircraft}: Same as above, we set each meta-learning episode to 20-way 5-shot classification with 15 query images per class. For each episode in PN models, in addition to calculating $L_{fewshot}$ using predicted pose heatmaps on FGVC, we also randomly sample 400 images from OID to calculate $L_{pose}$. In table \ref{tab:fgvc_hyper} we give all hyper-parameters chosen for training. All SGD optimizers use a default 0.9 momentum. For PN models, we select $\alpha{=}50$. During the whole training process, we evaluate the model on validation set every 40 epochs and select the best one to make the final evaluation on test set.

\begin{table}[h]
\centering
\resizebox{.85\textwidth}{!}{
\scriptsize
\setlength\tabcolsep{1.5pt}
\hskip-.02\textwidth
\begin{tabular} {  l  c c c c c c | c c c c c c}
\hline
\-\ & \multicolumn{6}{c}{\textbf{4-layer ConvNet}} & \multicolumn{6}{c}{\textbf{ResNet18}} \\
\hline
\textbf{Model} & \textbf{optimizer} & \textbf{lr} & \textbf{$\gamma$} &  \textbf{epoch} & \textbf{stage} & \textbf{weight decay} & \textbf{optimizer} & \textbf{lr} & \textbf{$\gamma$} &  \textbf{epoch} & \textbf{stage} & \textbf{weight decay} \\
\hline
proto & SGD & 0.1 & 0.1 & 500 & 2 & 1e-3 & SGD & 0.1 & 0.1 & 300 & 2 & 1e-3 \\
proto+PN & SGD & 0.1 & 0.1 & 500 & 2 & 1e-3 & SGD & 0.1 & 0.1 & 300 & 2 & 5e-3 \\
\hline
\end{tabular}
}
\vspace{.8mm}
\caption{The hyper-parameters selected for training on FGVC-Aircraft.}
\label{tab:fgvc_hyper}
\end{table}

\section{Numerical Experiment Results}

In table \ref{tab:cub_all}, we list all the numerical results for prototypical and dynamic based methods in all three evaluation settings.

\begin{table}[h]
\centering
\resizebox{.67\textwidth}{!}{
\scriptsize
\setlength\tabcolsep{5pt}
\hskip-.02\textwidth
\begin{tabular} {  l  c c c c c c}
\hline
\-\ & \multicolumn{3}{c}{\textbf{4-layer ConvNet}} & \multicolumn{3}{c}{\textbf{ResNet18}} \\
\textbf{Model} & \textbf{1-shot} & \textbf{5-shot} & \textbf{all-shot} &  \textbf{1-shot} & \textbf{5-shot} & \textbf{all-shot}\\
\hline
proto* & 14.63 & 27.63 & 32.09  & 23.77 & 38.76 & 42.73  \\
proto+MT & 17.05 & 31.52 & 35.56 & 30.90 & 47.78 & 50.93  \\
proto+BP & 15.07 & 28.36 & 35.56  & 21.04 & 37.15 & 41.04  \\
proto+FSL & 18.01 & 34.44 & 39.60  & 26.04 & 42.35 & 47.43  \\
proto+bbN & 15.87 & 30.63 & 37.75  & 24.05 & 39.60 & 44.02  \\
proto+uPN* & 19.06 & 39.48 & 46.24  & 28.06 & 47.10 & 53.18  \\
\textbf{proto+PN}* & \textbf{19.69} & \textbf{43.05}  & \textbf{49.56}  & \textbf{34.90} & \textbf{58.64} &\textbf{63.44} \\
proto+PN\_gt & 22.14 & 51.62 & 59.55 & 31.04 & 57.16 & 62.63  \\
\hline
dynamic & 16.02 & 28.59 & 35.77  & 24.24 & 38.64 & 43.27  \\
\textbf{dynamic+PN} & \textbf{27.77} & \textbf{47.72}  & \textbf{54.17}  & \textbf{33.98} & \textbf{54.27} &\textbf{60.19} \\
dynamic+PN\_gt & 34.00 & 56.32 & 62.67  & 33.90 &  54.60 & 60.09  \\
\hline
\end{tabular}
}
\vspace{.8mm}
\caption{All the numerical results on CUB. * indicates the results are averaged within 8 random trails for that model.}
\label{tab:cub_all}
\end{table}

\section{Training with Less Part Annotation}

Here we give the training details for the experiments in section \ref{ablation}. The training settings and hyper-parameters are the same as described in section \ref{details}. Similar to training on FGVC-Aircraft, we calculate $L_{fewshot}$ over each batch using the predicted pose heatmap. At the same time, we randomly sample a subset of images with part annotations from the same 20 classes to calculate $L_{pose}$. Here we define the batch size as the number of images with part annotation per class in each iteration. The detailed numerical results are shown in table \ref{tab:few_annot}.

\begin{table}[h]
\centering
\resizebox{.97\textwidth}{!}{
\scriptsize
\setlength\tabcolsep{5pt}
\hskip-.02\textwidth
\begin{tabular} {  l  c c c c c c c c c c c}
\hline
\-\ & \multicolumn{11}{c}{\textbf{percentage of training images with part annotation}}\\
 & 5 & 10 & 20 & 30 & 40 & 50 & 60 & 70 & 80 & 90 & 100 \\
\hline
batch size & 1 & 5 & 7 & 10 & 15 & 15 & 15 & 15 & 17 & 20 & 20 \\
\hline
4-layer ConvNet & 40.66 & 44.49 & 46.47 & 47.77 & 47.10 & 47.60 & 50.97 & 49.33 & 49.41 & 51.26 & 50.51 \\
ResNet18 & 52.78 & 54.17 & 57.53 & 61.41 & 63.51 & 64.52 & 64.35 & 65.57 & 65.87 & 65.99 & 66.33 \\
\hline
\end{tabular}
}
\vspace{.8mm}
\caption{For different percentages of training images with part annotation, we list the number of sampled part-annotated images per class per batch, as well as the final all-shot evaluation accuracy on both shallow and deep network backbones.}
\label{tab:few_annot}
\end{table}

\section{More Examples for Nearest Neighbor}
\begin{figure}[h]
\centering
\includegraphics[width=0.8\textwidth]{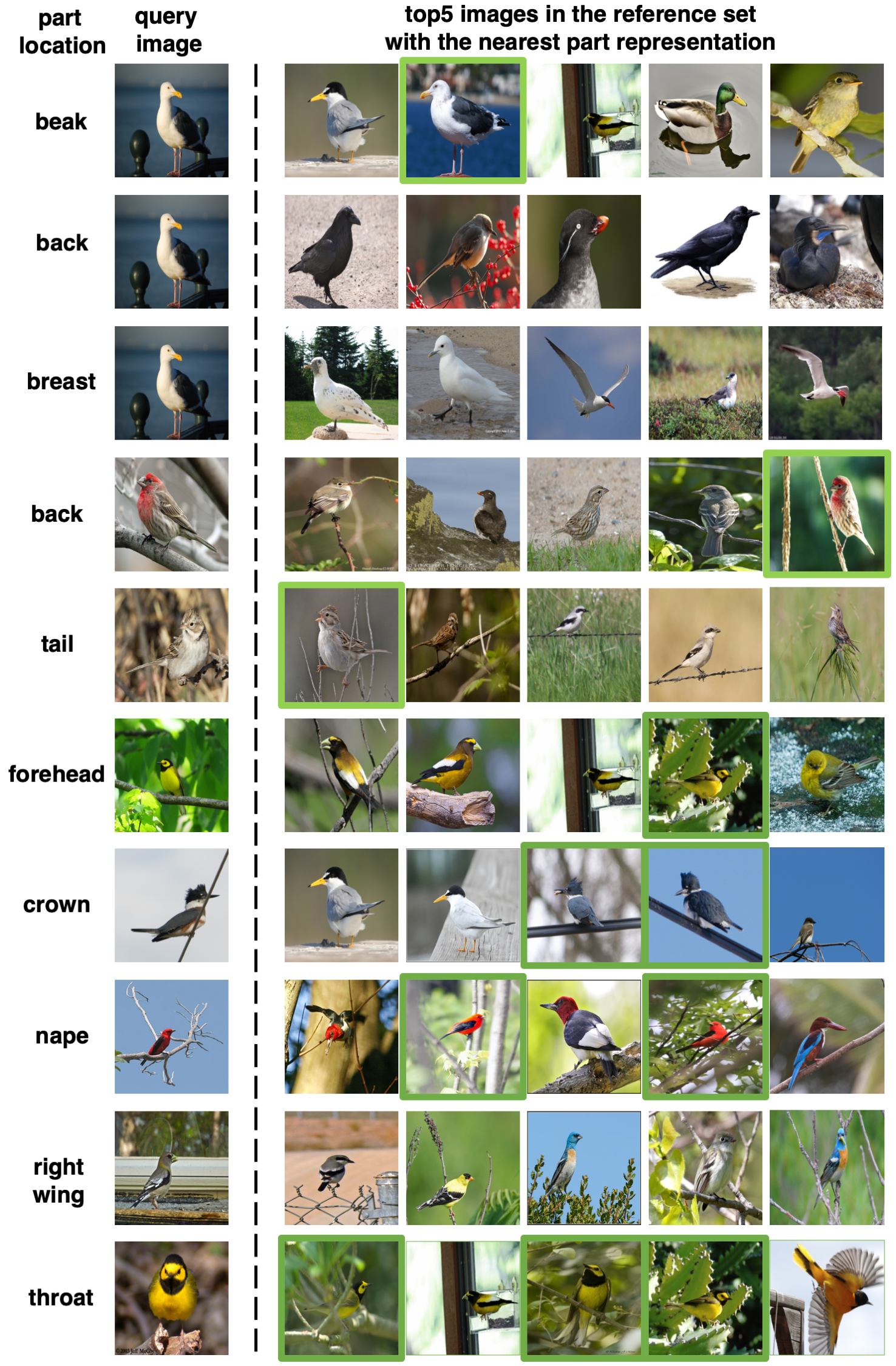}
\caption{Images with the closest part vector to the query image, for a given part location. Image is labeled with a green box if it belongs to the same class as the query image.}
\label{fig:neighbor_sup}
\end{figure}

In figure \ref{fig:neighbor_sup}, we show more visualization examples for the nearest neighbor experiment in section \ref{interpretation}.

\end{document}